\documentclass[lettersize,journal]{IEEEtran}
\usepackage{amsmath,amsfonts}
\usepackage{algorithmic}
\usepackage{algorithm}
\usepackage{array}
\usepackage[caption=false,font=normalsize,labelfont=sf,textfont=sf]{subfig}
\usepackage{textcomp}
\usepackage{stfloats}
\usepackage{hyperref}
\usepackage{url}
\usepackage{verbatim}
\usepackage{graphicx}
\usepackage{cite}
\newcommand{\p}[1]{\medskip \noindent \textbf{{#1}.}}
\newcommand{\eq}[1]{Equation~(\ref{eq:#1})}
\newcommand{\fig}[1]{Figure~\ref{fig:#1}}

\newcommand{\name}[1]{TransMASK}

\begin{document}

\title{TransMASK: Masked State Representation through \\ Learned Transformation}

\author{Sagar Parekh$^\ast$, Preston Culbertson$^\dagger$, and Dylan P. Losey$^\ast$
\thanks{$\ast$ Members of the Collaborative Robotics Lab (\href{https://collab.me.vt.edu/}{Collab}), Dept. of Mechanical Engineering, Virginia Tech}
\thanks{$\dagger$ Member of the Praxis Lab,
    Computer Science Department, Cornell University}
\thanks{Author's email: \href{mailto:sagarp@vt.edu}{sagarp@vt.edu}}
}



\maketitle

\begin{abstract}
When humans learn new manipulation skills, they are able to generalize these skills to new contexts and environments. In particular, when learning, humans can easily separate task-relevant aspects of the environment (e.g., object location) and distractors (e.g., the table color). Ideally, robot policies should be able to learn similar, generalizable representations, but instead they often fail under small environment shifts such as changes in lighting, object instance, or initial configuration. In this paper, we propose a self-supervised method that learns a \textit{mask} which, when multiplied by the observed image features, attempts to \textit{transform} these features to retain only those which are relevant to the task. Our method --- which we call \textit{TransMASK} --- can be combined with a variety of imitation learning frameworks (such as diffusion policies) without any additional labels or alterations to the loss function. By introducing a learned mask to the network during training, we aim to induce competitive pressure among the image features during training to force the policy to only attend to features which are consistently task-relevant. We find empirically that our masks are interpretable and can reject known spurious features such as the position of distractor objects or the background color. When compared to other representation learning methods for imitation learning, we find that TransMASK results in policies that are more robust to distribution shifts for irrelevant features, achieving at least $30 \%$ improvement over the baselines when tested on out-of-distribution environments. See our project website: \url{https://transmask.github.io/TransMASK/}

\end{abstract}


\section{Introduction}

\begin{figure*}
    \centering
    \includegraphics[width=0.9\linewidth]{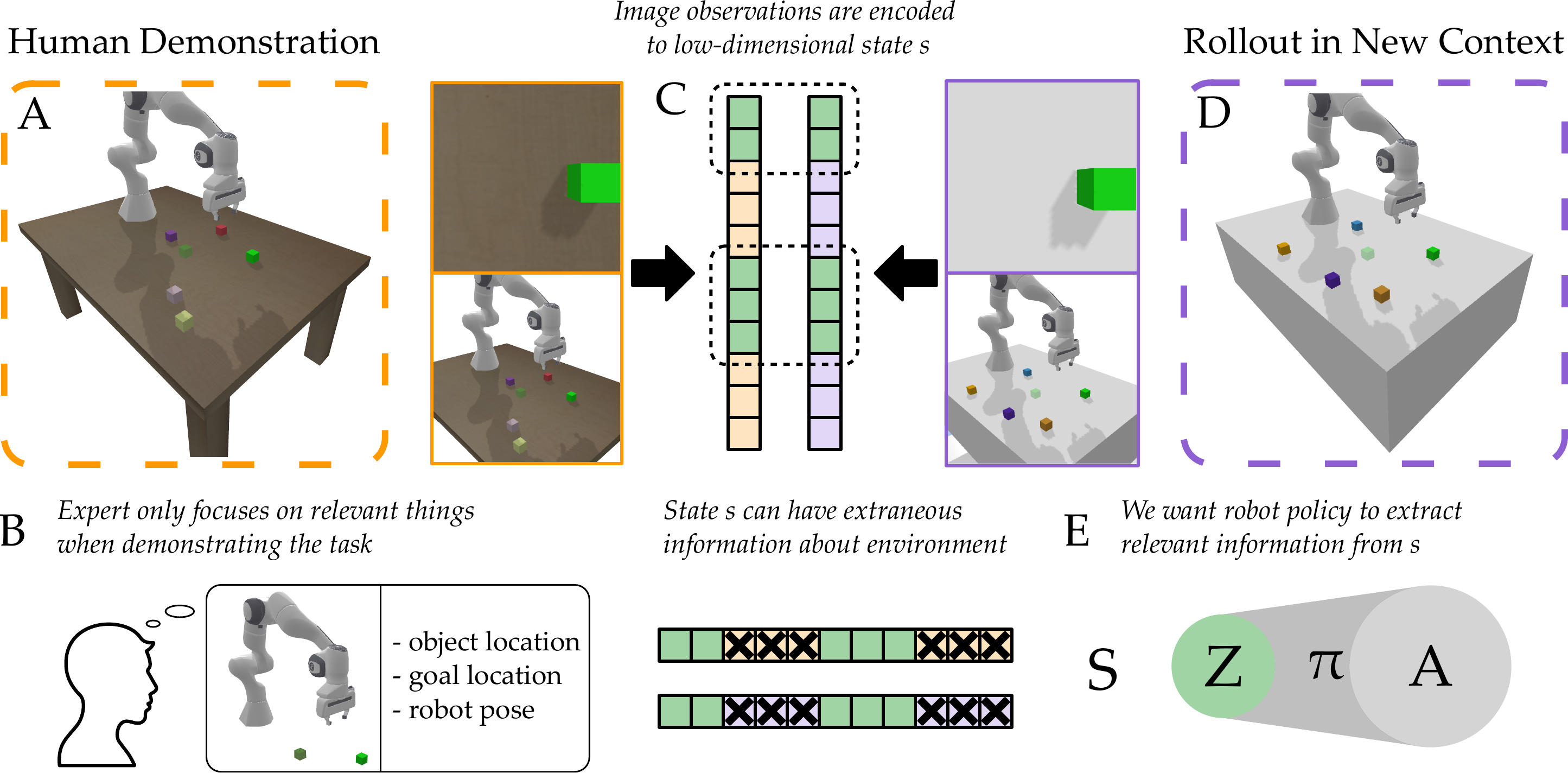}
    \caption{Robot learning to pick up a green block from a cluttered environment and place it at the center of the table. (A) The expert demonstrations are collected over a wooden table. (B) This expert's decisions depend only on features intrinsic to the task (e.g., green block, robot position, and target position). The robot's observations in these demonstrations, however, record information about the entire scene, capturing the table texture, background color, and task-irrelevant objects. (C) We assume a disentangled state $s$ is extracted from these high-dimensional visual observations. In this state, some elements correspond to the task structure --- positions of the green block, target, and robot, while others correspond to scene-specific factors. In the figure, relevant elements are shown in green and scene-specific elements are shown in orange and purple. (D) Standard imitation learning policy attends to the entire state. When this policy is deployed for the identical task in a different scene (e.g., over a marble table), it may fail due to spurious dependencies learned during training. (E) To learn a robust policy, we must encode the state to a representation $z$ which masks-out task-irrelevant features. A policy that is conditioned on $z$ is more robust to distribution shift caused by changes in the scene that do not alter the task structure, for instance, when the task is still to pick and place a green block but over a marble table instead of a wooden table.}
    \vspace{-1em}
    \label{fig:front}
\end{figure*}

Imitation learning enables robots to learn tasks from offline demonstrations provided by a human expert.
Consider teaching a robot to pick up a block and place it at the center of the table (see \fig{front}).
The human teacher guides the robot through successive states such as reaching the block, grasping it, and placing it at the desired location.
In the demonstration, the robot receives observations of the scene recorded through cameras and the actions applied by the expert.
When demonstrating the task, the human expert focuses only features crucial for completing the task --- the object, the goal, and the robot positions.
However, the robot's observations record information about the entire scene, capturing lighting conditions, table texture, background objects, and other task-irrelevant features.

A policy trained using these observations inherits this issue: it can attend to all parts of the state, including the details a human demonstrator would ignore. 
This causes the policy to become increasingly brittle.
Consider \fig{front}: when deploying a policy trained with demonstrations over a wood table for an identical task on a marble table, the visual observations become out-of-distribution, causing the policy to fail. 

Our key idea is that a robust policy is one that attends to the same features a human expert would. 
Thus, we aim to design policies with structures to mask out task-irrelevant information, enabling the policy to make decisions using features inherent to the task (e.g., object position, end-effector pose) instead of the scene(s) in its training data.

Existing works have developed a variety of methods for extracting this relevant state.
A straightforward approach augments the training data with random transformations of visual observations (e.g., random cropping, adding noise to the image) \cite{tobin2017domain,kaidanov2024role}.
Nevertheless, introducing variability in the training data can degrade in-domain performance and does not guarantee robustness to large distribution shifts.
Another line of work leverages vision-language model (VLM) backbones \cite{sharma2023lossless,yang2024robot} that have robust state representations due to large-scale pretraining on diverse datasets.
Despite these advantages, VLMs typically require task-specific finetuning, which can lead to catastrophic forgetting of pretrained features adversely effecting the robustness.
Alternatively, a different class of literature focuses on learning minimal state representations that can extract only the most relevant information from the observations.
This is achieved by introducing a trade-off between i) disconnecting the latent state from the actual state (compression) and ii) ensuring the latent state can map to correct actions (performance).
Examples of this approach include using variational autoencoders (VAEs) for encoding the images to states \cite{rahmatizadeh2018vision,shang2021self}, training an encoder with the Information-Bottleneck (IB) principle \cite{tishby2000information}, or adding structure to the representation space using contrastive learning \cite{lee2025class,giammarino2025visually}.
However, these approaches --- as we will show empirically and theoretically --- rely on an ill-posed optimization problem that often collapses into inaccurate state representations that lack either compression or performance.

We therefore propose a new method for extracting the correct state from observations without modifying the optimization objective.
We begin with the intuition that actions demonstrated by the expert depend only on parts of the state that are relevant to the task.
Although nuisance variables may also be correlated with the demonstrated action, variation in these variables that does not affect the expert action provides a signal for distinguishing them from task-relevant state. Building on this intuition, we hypothesize that:
\begin{center}
    \textit{Imitation gradients can provide a useful signal for identifying the task-relevant state.}
\end{center}
Put simply, by introducing competitive pressure between features, we can exploit the gradients to identify and preserve the components of state that matter for control.
This prevents the robot policy from learning spurious correlations between states and actions.
In order to ensure that the state space is structured, we assume that the robot state is disentangled and can be decomposed into relevant and irrelevant elements.
For example, the first $k$ elements of the state could represent the location of the block and the robot\footnote{We clarify that obtaining such disentangled features for our method does not require additional domain knowledge. As we will show later in our experiments, segmentation masks for all objects present in the scene can satisfy this assumption without the need for additional information from the designer, since the object labels can be simply obtained from a pretrained vision language model without needing finetuning.}.
Despite the additional structure imposed on the state space, the individual states can still contain environmental noise which can lead to causal confusion.
In our running example, if the environment has other blocks in addition to the green one, the policy can incorrectly map the position of these ``distractors" to the action.
Consequently, a change in the position of any of these extra blocks can cause the state to go out-of-distribution even when the green block is not moved.

Applying this insight we derive \name{}, a method that can extract relevant information from the state without additional supervision.
\name{} relies on the underlying optimization scheme of imitation learning and utilizes its gradients to learn a transformation matrix.
This matrix transforms the robot's state into a representation which preserves elements important for action prediction while suppressing the rest.
Our approach is a step towards robots that learn robust policies from human demonstrations, and can apply their policies to new testing environments.

Overall, we make the following contributions:


\p{Finding Limitations of Existing Approaches}
We show how existing approaches to state representation are not likely to the invariance required in our imitation learning setting.
Information-bottleneck methods suppress nuisances through global compression, which also penalizes task-relevant information and requires selecting an appropriate bottleneck strength.
Conversely, reconstruction-based approaches encourage the latent to preserve nuisance variation along with the relevant state.
These limitations motivate a more structured bottleneck over state features, using the imitation objective to determine which features should be retained.

\p{Deriving \name{}}
We specifically focus on settings where the state is disentangled into relevant and irrelevant elements.
Within this setting we develop a state representation method that naturally learns a representation $z$ of the relevant state while purely optimizing for performance.
Our approach is linked to attention networks, and we harness the properties of the policy Jacobian to reach our proposed approach.

\p{Comparing to Baselines}
We compare our approach to state-of-the-art representation baselines across a variety of simulated and real-world environments with visual as well as privileged state information.
\name{} achieves up to $15 \%$ higher performance than the next best baseline when tested in settings without distribution shift and approximately $9 \%$ higher success under environmental perturbations.

\section{Related Works}

\noindent \textbf{Robust Imitation Learning.}
Imitation learning has advanced rapidly in recent years, driven by the development of expressive architectures such as diffusion policies \cite{chi2025diffusion} and transformer-based policies \cite{rt22023arxiv}. 
Despite these successes, deploying imitation learning policies in scenarios that differ from the training distribution (e.g., different object locations, clutter, or visual features) remains an open challenge \cite{gupta2025adapting,mao2025omnid}.
Several works have attempted to address this challenge and make policies robust to distribution shift, out-of-distribution observations, or noisy observations.
Broadly, these works focus on two distinct approaches to achieve robustness: one set of methods focus on the \textit{training data} while another focuses on the \textit{training procedure}.

\textit{Data curation and diversification}:
The first class of methods focus on structuring the demonstration dataset so that the learned policy attends to the most relevant aspects of the state space.
A straightforward strategy is to collect datasets while ensuring data diversity to improve robustness \cite{saxena2025matters,shi2025diversity}.
However, in many practical settings, we may have limited control over data collection, since it is common to use readily available datasets such as \cite{10611477,robomimic2021}.
Therefore, several approaches aim to re-balance the given dataset to learn diverse behaviors reliably \cite{parekh2025towards,hejna2024re}.
Other works improve generalization through domain randomization \cite{tobin2017domain,kaidanov2024role}, exposing the policy to variations in the observations during training.
Alternatively, we can augment the available dataset with auxiliary data to improve robustness of the trained policy.
For example, \cite{xu2022discriminator,kim2021demodice} leverage auxiliary datasets alongside optimal data to improve performance, while \cite{ke2023ccil,park2022robust} learn a dynamics model from expert data and use it to generate additional synthetic demonstrations.

\textit{Adversarial BC}: 
The second class of methods modifies the training pipeline to enhance robustness.
This is typically achieved through adversarial training which improves model robustness by minimizing the prediction loss under worst-case perturbations of the input.
In the context of imitation learning this is formulated as a generative adversarial game in which the policy is trained jointly with a discriminator that distinguishes between expert trajectories and those generated by the policy.
The most prominent works in this domain include GAIL \cite{ho2016generative} and f-GAIL \cite{zhang2020f}, while other works have extended this idea to achieve better sample efficiency and performance \cite{wang2022adversarially,jung2024sample}.

While these approaches in imitation learning have been shown to improve performance and robustness, they do not address the more fundamental problem: we should condition the policy only on the relevant parts of the state.
Data curation approaches can have computational overheads related to generating additional data or finding the right balance for the demonstration's prior to training the policy. 
Some of these methods rely on a human expert for labeling the dataset which can be time consuming.
Adversarial training can often introduce instability in optimization, requiring precisely tuning the discriminator, and is often unsuitable for high-dimensional inputs such as visual observations.
Additionally, variants of GAIL are sample-inefficient as they handle distributional shift by interacting with the environment. 

\p{State Representations}
There are various works that explicitly focus on extracting relevant information about the task to achieve robust performance by encoding the observations to a structured representation space.
Recent advances in object-centric representations, such as VIMA\cite{jiang2023vima} and Transporter Nets \cite{zeng2021transporter}, decompose the scene into object tokens using segmentation.
While this has shown to improve spatial reasoning, they attend to all discovered entities equally.
A more general approach requires compressing the input by retaining only critical information from the state and discarding redundant or unnecessary features.
Contrastive learning, which optimizes the latent representation space by minimizing the distance between semantically similar samples while maximizing the distance between dissimilar samples, has emerged as a viable solution to this problem.
In imitation learning, contrastive loss is leveraged to learn representations that are robust to variations in visual observations \cite{lee2025class,yang2025invariance}.
While \cite{yang2025invariance} uses auxiliary static observations of the robot alongside demonstrations and identify positive and negative pairs using temporal proximity of the observations, \cite{lee2025class} uses action sequence similarity to identify positive and negative pairs. 

However, a contrastive approach typically requires comparing each representation with many negative examples to prevent representation collapse.
An alternative is to use the Bootstrap Your Own Latent (BYOL) objective \cite{grill2020bootstrap} to learn invariant state representation from image observations as in \cite{pari2021surprising}.
A simpler approach is to leverage self-supervised training using variational autoencoder (VAE) for extracting representations.
For example, \cite{xihan2022skill,noseworthy2020task} use VAE to extract motion primitives from robot trajectories; and \cite{rahmatizadeh2018vision,park2021object} use VAE to extract better state representations from image observations.

Learning state representation has also been viewed from an information theoretic perspective, where mutual information can eliminate redundancy in the learned representations.
Specifically, research has examined the use of the Information Bottleneck (IB) principle \cite{tishby2000information} for deep representation learning, demonstrating that IB can make latent representations invariant to irrelevant parts of the inputs \cite{achille2018emergence,achille2018separation}.
This principle has also been leveraged for learning task-relevant state representations for robust imitation learning \cite{bai2025rethinking}.

Despite their success, there are limitations to these approaches.
First, contrastive learning requires identifying positive and negative samples in the demonstrations which requires additional domain knowledge.
Training can also be unstable in contrastive learning resulting in representation collapse.
Further, approaches like VAEs can encode high-dimensional visual observations to a structured representation space, but they do not explicitly aim to retain only task-relevant information.
Finally, while the IB seeks to compress the state, in the context of imitation learning, it often lacks an explicit supervision signal to distinguish between nuisance factors and features important for task completion.
Later in the text, we will provide a detailed discussion on why the IB objective is ill-posed for imitation learning applications.
In contrast, \name{} extracts task-relevant features as a byproduct of the imitation learning gradient flow, alleviating the need for additional data or modifying the training pipeline and the objective function for optimization.

\section{Problem Statement} \label{sec:problem}

We consider offline imitation learning settings.
Within these settings, the robot is given a fixed dataset of demonstrations.
In each demonstration a human expert teleoperates the robot through instances of the desired task.
Based on these examples the robot learns a control policy that can robustly perform the demonstrated task across a variety of initial states.
In particular, the learned control policy should be robust to changes in the scene that do not alter the task (e.g., if the color of a table changes, or if clutter is added to the background).
We specifically focus on settings where the state is \textit{disentangled}.
There are some elements of the state that are relevant to the expert's control decisions, while other state elements are irrelevant noise that do not impact how the expert behaves.
To separate the relevant and irrelevant state elements, we introduce a latent state representation $z \in \mathbb{R}^d$.

\p{Dataset}
Each demonstration consists of a sequence of tuples $\xi = \{(x^1, I^1, a^1), \cdots , (x^t, I^t, a^t)\}$ where $x \in \mathbb{R}^m$ is the proprioceptive state (e.g., joint angles), $a \in \mathbb{R}^m$ is the applied action (e.g., joint velocities), and $I$ is the robot's observation (e.g., images captured from onboard and/or static camera(s)).
Overall, the dataset includes a few demonstrations for the desired task.
In our settings, we consider state $s \in \mathbb{R}^n$ which includes the proprioceptive states $x$ as well as features $\phi$ encoded from image observations $I$, i.e., $s = \left[ x, \Phi(I) \right]$ using visual encoder $\Phi$.
For every state in the dataset, the human expert labels what action the robot should take in order to complete their desired task.

\p{Human}
The expert's policy $\pi^*$, which they use to provide demonstrations, only considers task-relevant features and is not influenced by scene-specific components.
For example, when demonstrating how to pick up a block, the human only focuses on the block's location and the robot's location relative to the block.
Their decisions are not influenced by texture of the table, lighting conditions, or the presence of clutter\footnote{Provided that such clutter does not interfere with grasp feasibility. In our setting, clutter is treated as a potential source of spurious correlation or causal confusion, rather than an obstruction to manipulation.}.
However, the state $s$ that the robot records when the human is demonstrating the task includes all the information from the scene.


\p{Disentangled State}
In our problem setting, we assume that the state vector $s$ is disentangled and can be separated into task-relevant components $\mu$ and nuisance variables $\eta$. Without loss of generality, we write $$s = (\mu, \eta),$$ where this notation reflects a reordering of the variables, and does not imply that we know the correct partition \emph{a priori}.
Here $\mu \in \mathbb{R}^k$ are the \textit{relevant} elements of the state that the human expert uses to make their control decisions.
Returning to our running example, $\mu$ could contain the location of the block.
The remaining state elements $\eta \in \mathbb{R}^{n-k}$ are \textit{irrelevant} to the expert's control decisions, and can be treated as extraneous information.
For example, perhaps $\eta$ includes the color of the table or the location of background objects.

We emphasize that the \textit{disentanglement} of state $s$ is fundamental to our proposed approach.
We argue that this sort of separation is often inherent to the learning context (e.g., humans only focus on a few relevant features when making decisions \cite{bera2021gaze}).
However, we recognize that the state $s$ that the robot extracts from the scene may not align with these disentangled representations, and we will test the limits of this assumption in our experiments.

\p{Latent State}
Since the state $s$ can contain irrelevant information, we do not want our robot learner to make decisions based on every aspect of this state.
To make the policy robust, it should focus only on task-specific information $\mu$, but this partition is unknown to the learner. We assume that the expert policy is invariant to the nuisance variables,
\begin{equation*}
\pi^*(\mu,\eta)=\pi^*(\mu,\eta^\prime)
\end{equation*}
for all admissible $\eta,\eta^\prime$.

We seek a learned representation $z=f(s)$ that inherits this invariance while retaining the information needed to reproduce the expert action. Ideally,
\begin{equation}
f(\mu,\eta)=f(\mu,\eta^\prime)
\label{eq:rep_invariance}
\end{equation}
for all admissible $\eta,\eta^\prime$, while there exists a policy $\pi$ such that
\begin{equation}
\pi^*(s)=\pi\left(f(s)\right).
\label{eq:policy_performance}
\end{equation}
desired representation is insensitive to nuisance variation without discarding information required for the task.

\p{Policy}
Using the offline dataset $\mathcal{D}$ we jointly learn the state mapping $f_\theta$, and a control policy $\pi_\psi$ that automates the demonstrated task, parameterized by $\theta$ and $\psi$ respectively.
We write our overall policy as $a = \pi_\psi\big(f_\theta(s)\big)$.
Our objective is for this compositional policy to achieve two things.
To learn performant behaviors, the actions $a = \pi(z)$ should match the expert's demonstrated actions across dataset $\mathcal{D}$.
To increase robustness to environmental variations, the latent $z = f_\theta(s)$ should remain invariant to $\eta$.
\section{TransMASK}

We want the robot to learn tasks from human demonstration while being robust to environment variations.
As previously discussed, only the task relevant features are factored into the expert's decision making, and a robust robot policy should only be conditioned on these aspects of the state $\mu$.
To extract $\mu$, we propose a method to learn state representations $z = f(s)$ that fulfill two desired characteristics --- it must retain information from $s$ that enables the policy to match the actions demonstrated by the human, while removing information $\eta$ that does not relate to the task structure.
In this section, we start by demonstrating how standard approaches for learning latent representations --- such as autoencoders --- are poorly designed for this objective (Section~\ref{sec:M1}).
Next, we analyze the Jacobian of the expert policy to determine the characteristics of an effective state representation (Section~\ref{sec:M2}).
Finally, we derive our proposed framework for learning $z$, and prove that this approach addresses the issues with standard representation techniques (Section~\ref{sec:M3}).

\subsection{State Representation and Collapse} \label{sec:M1}
Equations~(\ref{eq:rep_invariance})--(\ref{eq:policy_performance}) describe two desired properties of the latent state. The representation should be invariant to nuisance variation while retaining enough information to reproduce the expert action. These conditions are difficult to impose directly because the learner does not know which components of the observed state belong to $\mu$ or $\eta$.

We can approach this problem through an information-theoretic lens by posing it as an optimization. If we knew the proper partition of $s$, we would see a representation of the state that maximizes mutual information between the latent and the expert action, while containing little additional information about the nuisance variables, conditioned on the relevant state:
\begin{align*}
    \max_f  & \qquad \mathcal{I}(z; a) \\
    \text{s.t.} & \qquad \mathcal{I}(\eta; z \mid \mu) \leq \epsilon,
\end{align*}
where $z = f(s),$\footnote{For deterministic encoders of continuous variables, $\mathcal{I}(s;z)$ is potentially infinite. For the purposes of this discussion, all MI terms in this section consider $\tilde{z} = f(s) + \zeta$, for $\zeta \sim \mathcal{N}(0, \sigma^2 I)$, with the tilde suppressed for readability.} and $\mathcal{I}(A;B \mid C)$ is the conditional mutual information between $A$ and $B$, conditioned on $C$. We can form the Lagrangian of this optimization, for a fixed $\lambda \ge 0$, which provides a desired loss
\begin{equation}
    L_\text{des} = \lambda \mathcal{I}(\eta; z \mid \mu) - \mathcal{I}(z; a),
    \label{eq:lagrangian}
\end{equation}
up to a constant. 

A related approach to learning robust state representations is given by the Information Bottleneck (IB) principle~\cite{tishby2000information}. IB replaces the conditional MI in Equation~\eqref{eq:lagrangian} with a general state compression objective:
\begin{equation}
    L_\text{IB} = \lambda \mathcal{I}(s;z) - \mathcal{I}(z;a).
    \label{eq:ib}
\end{equation}
Intuitively, this loss trades off compression (i.e., minimizing MI between the state $s$ and the representation $z$) and performance (i.e., maximizing the MI between the latent $z$ and the expert action $a$). We can compare the two objectives using the partition of the state, $s = (\mu, \eta)$ and the chain rule of mutual information, $\mathcal{I}(s;z) = \mathcal{I}(\mu;z) + \mathcal{I}(\eta ; z \mid \mu),$
which, substituting into Equation~\eqref{eq:ib}, yields the expression:
\begin{align*}
    L_\text{IB} &= \lambda \mathcal{I}(\eta;z \mid \mu) - \mathcal{I}(z;a) + \lambda \mathcal{I}(\mu;z).
\end{align*}
The first two terms align with our desired objective by encouraging a representation that predicts the expert action while suppressing nuisance-specific variation. However, the additional term $\lambda \mathcal{I}(\mu;z)$ also penalizes information retained about the task-relevant state. Standard IB therefore suppresses the nuisance indirectly by compressing the entire state rather than specifically targeting nuisance variation.

While the IB loss does not require knowing the correct partition of the state, minimizing~\eqref{eq:ib} explicitly can be quite difficult. IB formulations typically choose a fixed value of the Lagrange multiplier $\lambda$, which forces a particular and difficult-to-tune tradeoff. When $\lambda$ is too small, there is little incentive to compress the state, which means that the representation may retain nuisance signals which remain predictive on the training distribution. When $\lambda$ is too large, the compression objective dominates, and the representation can collapse and lose information which is required to predict the expert's action. In general, the correct value of the parameter is task- and context-dependent, and it usually must be selected carefully for each application. In our BC setting, the tradeoff in the IB loss is also somewhat unnecessary, since our goal is not to compress the state, but instead to suppress the elements of the state which are nuisance variables.

An alternative is to explicitly encourage the latent to retain state information, for example by reconstructing $s$ from $z$ as in an autoencoder. This avoids the need to determine how much state information should be discarded, but introduces the opposite bias. Reconstruction losses encourage $z$ to preserve both task-relevant variation in $\mu$ and the nuisance variation in $\eta$. Thus, neither generic compression nor full state reconstruction directly suppresses the nuisance inputs $\eta$.

Another key issue with IB formulations is that they can be difficult to directly optimize in high-dimensional, continuous settings. Mutual information is often difficult to compute in closed form, and practical estimators require variational bounds or approximate $\mathcal{I}(s;z)$ using a KL penalty from a fixed prior. While tractable, these methods often impose additional structure on the latent state, limiting the expressiveness of the resulting representation.

These issues, along with our assumption of a disentangled state, motivate a more structured form of bottlenecking which requires a less explicit tradeoff between compression and performance. Rather than explicitly minimizing the IB using a particular choise of $\lambda$, we instead  propose to bottleneck the state using a learned mask, which uses competition between different features to implicitly suppress redundant or noisy nuisance features. This formulation allows us to preserve the minimality intuition of the IB objective but without requiring knowledge of the correct partition, explicit estimation of MI, or introducing additional compression losses to the behavior cloning objective.

\begin{figure*}
    \centering
    \includegraphics[width=0.8\linewidth]{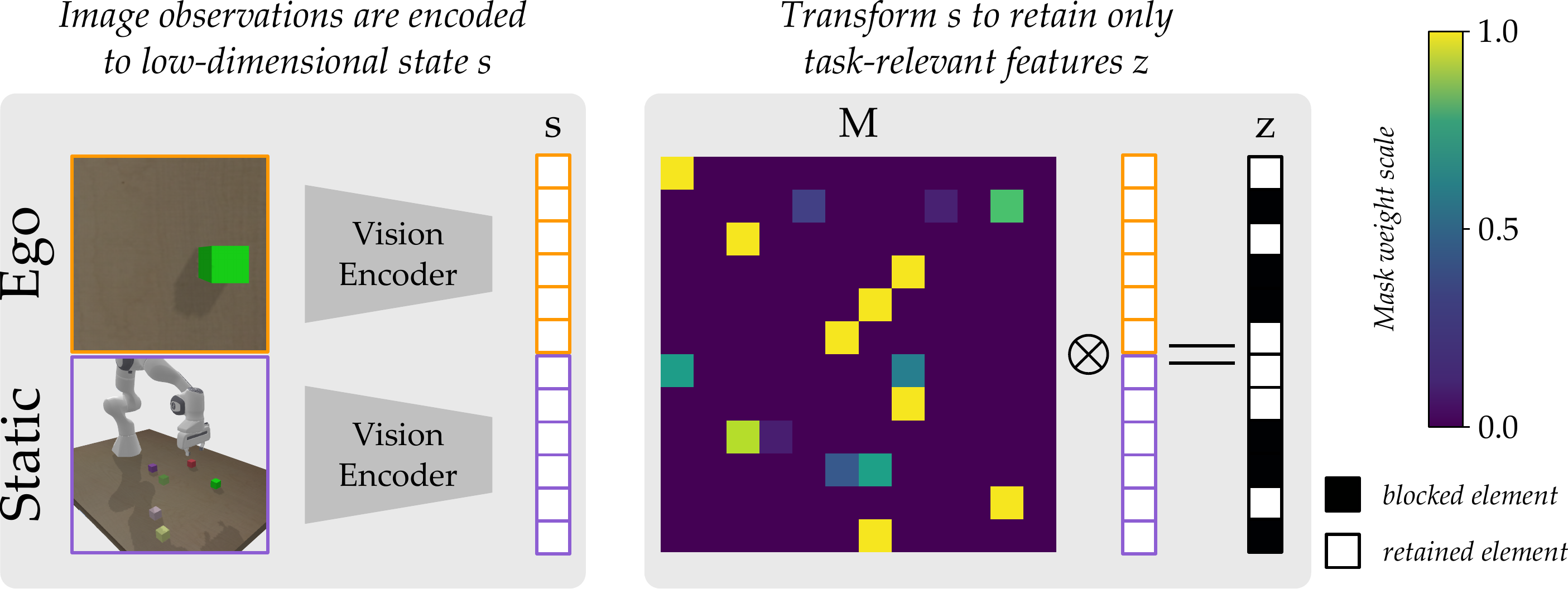}
    \caption{Schematic diagram of TransMASK. The visual observations are encoded to a disentangled state vector $s$. We introduce a mask encoder that outputs a constant mask $M$ of shape $n \times n$, where $n$ is the dimension of $s$. This mask is a sparse matrix in which the columns that correspond to task-irrelevant elements of $s$ are close to zero. Therefore, when we compute the representation $z$ by transforming the state with $M$, it only retains the elements critical for accurately predicting actions. The robot policy is conditioned on $z$ rather than the entire state to achieve robustness.}
    \label{fig:method}
\end{figure*}

\subsection{Jacobian of the Expert Policy} \label{sec:M2}
Instead of using the ill-posed optimization of the IB principle for imitation learning, we intend to optimize for performance alone and utilize the underlying gradient structure to extract a compact representation of the state.
We start by taking a closer look at the expert policy $\pi^*(s)$.
Remember, in our problem settings we assume that the state $s = (\mu, \eta)$ is composed of relevant and irrelevant elements, and these elements are disentangled, i.e., $\mu$ and $\eta$ are mutually exclusive and exhaustive ($\mu \cap \eta = \emptyset$, $\mu \cup \eta = s$).
With this in mind, consider the Jacobian of the expert policy $\nabla_s \pi^*(s)$. 
This Jacobian is a $m \times n$ matrix, where $m$ is the dimension of the action $a$ and $n$ is the dimension of the state $s$.
The columns of this matrix indicate the impact of the corresponding element of the state on the robot's actions, or in other words, the sensitivity of the action to the corresponding element of state.
For example, the first column indicates how the first element of $s$ affects each dimension of $a$.
Since the expert's decisions are not influenced by the irrelevant elements of $s$, the column $i$ of the Jacobian corresponding to $s_i \in \eta$ will be all zeros.
In contrast, the columns $j$ of the Jacobian corresponding to $s_j \in \mu$ will have non-zero values since these elements affect the expert's actions.
Overall, $\nabla_s \pi^*(s)$ will be a sparse matrix such that the states that contain purely irrelevant information are in the \textit{null space} of the Jacobian.
In the next subsection, we discuss how this motivates our design of a mask-based state representation, with the aim of matching the expert's causal structure.

\begin{figure}
    \centering
    \includegraphics[width=\linewidth]{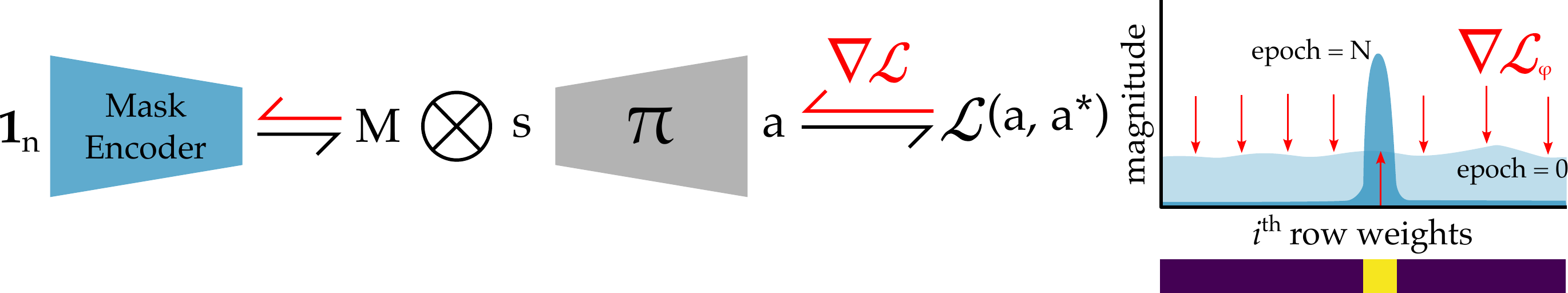}
    \caption{Overview of how the mask is learned. (Left) The mask encoder inputs a $1$-vector of size $k$ ($\textbf{1}_k = \left[ 1, 1, \cdots, 1 \right] \in \mathbb{R}^k$) and outputs a matrix $M \in \mathbb{R}^{n \times n}$. Since the expert's actions when providing demonstrations are only influenced by the task-relevant features in the environment, the columns of the Jacobian of the expert policy corresponding to the irrelevant elements in the state will be near-zero magnitude as discussed in Section \ref{sec:M2}. (Right) When training the robot policy, as the loss converges the Jacobian of the robot's learned policy changes the rows of $M$ until the values start weighting the task-relevant elements more than the irrelevant ones.}
    \label{fig:method_gradient}
\end{figure}

\subsection{Deriving \name{}} \label{sec:M3}
A practical difficulty with the standard IB formulation is that the desired amount of compression is not known a priori. Implementations therefore typically select a fixed tradeoff coefficient $\lambda$ to balance predictive information against compression. In our setting, however, we specify which variation should be ignored, rather than a desired level of state compression. We therefore avoid an explicit global compression objective and instead impose competition directly among state features through the architecture.

To this end, we define the state representation $z$ as an affine transformation of state, i.e. $z = Ms$, 
where $M$ is a $n \times n$ square matrix.
We will refer to $M$ as a \textit{mask}, since it can mask-out or remove elements of $s$ from the resulting $z$.
Indeed, each element of $z$ here is a linear combination of the state elements, i.e., $z_i = M_{i1}s_1 + \ldots + M_{in}s_n$.
Selecting $z = Ms$, where $M$ is a constant matrix, ensures that the state representation maintains a correlation with the state and does not collapse into an action representation.
This also eliminates the need for estimating the mutual information between $s$ and $z$, which is often intractable. 
Further, as we will empirically show later, our design of $M$ produces strong regularization without requiring an explicit loss term.
We train the model using a simple behavior-cloning MSE loss:
\begin{equation}
    \mathcal{L}(\psi, M) = \sum_{(s, a) \in \mathcal{D}} \frac{1}{2} \| \pi_\psi(M s) - a \|^2 \label{eq:loss}
\end{equation}
Since the matrix $M$ is constant, we parameterize it with a learnable parameter $\theta$.
The state $s$, which contains disentangled features obtained from image observations, is transformed into a representation $z = Ms$.
As a result, our approach is versatile because the mask can be easily appended to any existing IL policy without augmenting the training process or the loss function.
We show later in our experiments that this enables the network to learn a mask that only focuses on parts of the state that relate to the task structure $\mu$ while removing the unimportant elements $\eta$.
Our method is summarized in \fig{method}.

To demonstrate how the Jacobian can naturally learn such a mask, we take the gradient of our loss in \eq{loss} with respect to the mask weights $\theta$:
\begin{align}
    \frac{\partial \mathcal{L}}{\partial \theta} &= \left(\frac{\partial (\pi(z) - a)^T (\pi(z) - a)}{\partial M}\right)^T \frac{\partial M}{\partial \theta} \nonumber \\
    &= s \frac{\partial (\pi(z) - a)^T (\pi(z) - a)}{\partial z} \frac{\partial M}{\partial \theta}\nonumber \\
    &= s \cdot (\pi(z) - a)^T \cdot J_{\pi} z \cdot \frac{\partial M}{\partial \theta} \nonumber \label{eq:jacobian}
\end{align}
Here, we apply the vector chain rule through the computational graph, and $J_{\pi} z$ is the Jacobian of the policy with respect to z.
If $M$ is a learnable parameter matrix $\theta \in \mathbb{R}^{n \times n}$, this gradient of the loss is an $n \times n$ matrix.
On the other hand, if $M$ is the output of a neural network with parameters $\theta$, the gradient of the loss becomes a tensor of shape $n \times n \times p$, where $p$ is the input dimension of the network.

Because the expert policy is invariant to the nuisance inputs (i.e., $\frac{\partial \pi}{\partial s_i} =0$ for $i \in \eta$), our goal is to learn a policy that has the same causal structure. This motivates our choice of a learnable mask. As a simple example, consider the case of $s \in \mathbb{R}^2$, with $s = [s_1, s_1 + \epsilon]^\top$, for $\epsilon \sim \mathcal{N}(0, \sigma^2),$ and an expert policy where $\pi^*(s_1, s_2) = \pi^*(s_1, 0)$. Although, in this case, $s_2$ may be strongly correlated with $s_1$ and the demonstrated action, it provides no additional information to the expert when $s_1$ is available, and introduces additional noise (which may shift during test-time) to the policy's computation. Thus, we prefer a learned representation $z$ which assigns zero weight to $s_2$, avoiding unnecessary reliance on this potentially spurious correlation.

To encourage this behavior, we normalize each row of $M$, e.g. by applying softmax, over the state dimensions. This normalization creates competition among features because increasing the weight assigned to one feature necessarily decreases the weight assigned to others. Assigning mask weight to a nuisance feature therefore consumes a fixed weight budget to encode information that is redundant with task-relevant state, while also admitting nuisance-specific variation. The imitation objective can consequently favor allocating mask weight to features that more directly and reliably explain the expert action.

Empirically, we find that even without explicit sparsity regularization, this competitive pressure often produces highly concentrated masks. Individual rows tend to assign nearly all of their weight to task-relevant features while suppressing nuisance features. As a result, $M$ often acquires an approximately sparse structure in which columns corresponding to important state features $\mu$ retain substantial weight, whereas columns corresponding to nuisance variables $\eta$ are largely suppressed. The learned mask therefore acts as a feature selector that filters task-irrelevant variation while retaining the information required to reproduce the expert policy.


\subsection{Comparing with Attention Mechanism}
Here we provide an alternative perspective on our proposed \name{} framework.
Our mask can be viewed as an attention mechanism \cite{vaswani2017attention} where we learn the attention weights for the elements of the input state; the state representation is the matrix product of the state with these learned attention weights.
In attention mechanisms, the attention weights for the token pairs is defined by the formula: $\text{softmax}\left( \frac{QK^T}{\sqrt{d_k}} \right) V$.
During training, there is no direct supervision for learning the correct attention weights.
Rather, the back propagation of gradients from the loss push query $Q$ and key $K$ to produce a larger dot product for the important pairs and lower dot product for unimportant pairs.
As a result, the networks that learn the key, query, and value are updated such that the model is able to discover the correct attention patterns that minimize the loss.
Similar to this, in our method there is no direct supervision for training the mask.
Rather, it is learned as a result of the gradients that encourage the mask elements to focus more on the parts of $s$ which influence the output while ignoring the irrelevant parts.

Nevertheless, our method fundamentally differs from conventional attention mechanism in the way it weights the elements of the state.
Attention mechanisms are designed to model the relative importance of different components of the input, whether these correspond to different elements in a sequence or spatial patches of images.
In contrast, our task requires transforming a single disentangled state vector into a compressed representation that retains only the task-relevant elements.
At first glance, attention might appear to be a natural solution.
However, a standard attention mechanism computes dynamic attention weights that are conditioned on the input, i.e., the key, query, and value are computed via neural networks that take the state vector as input.
While this allows the model to focus adaptively on different elements of the state, it also introduces the risk of losing the original disentanglement as task-irrelevant elements may be inadvertently weighted and propagated through the representation.
In contrast, our method uses the policy Jacobian as supervision for feature selection.
As a result, we learn a static mask that is input-invariant as it is independent of the state.
This static mask reflects our underlying assumption about the state decomposition into $\mu$ and $\eta$.
This design is particularly advantageous because, even when deploying a policy in a new environment, task relevance should be determined by task structure, rather than the variations in the input values. 
For instance, in a pick-and-place task, the end-effector and object poses are always relevant, while visual textures, or background color are generally not relevant. 
Thus, making our attention mask input-dependent, as in the conventional attention mechanism, would simply provide another path to allow task-irrelevant information to enter the policy. 

To summarize, our method shares some similarity with the attention mechanism in the learning dynamics --- the way attention mechanism is able to learn the attention weights is closely related to how our mask learns which elements of the state to attend to.
However, it has a principal distinction from the attention mechanism.
\name{} learns a static, state independent mask leveraging the disentangled structure of the state, transforming it into a meaningful state representation for downstream policies.

\section{Experiments}

\name{} learns a mask that attends only to the elements of the state that are relevant for performing the given task.
In this section, our goal is to validate the proposed approach and test whether it can achieve robustness against data which is outside the training distribution.
We conduct extensive experiments in simulation as well as real-world tasks.
Additionally, we incorporate our method with two policy classes: MLP, which uses fully-connected architecture as the policy head; and Diffusion Policy (DP), which is a policy head trained with the diffusion objective \cite{chi2025diffusion}.

\p{Baselines} Across our experiments, we compare \name{} with state-of-the-art state representation baselines. 
These baselines learn well-structured representations for the state either through pretrained visual encoders, contrastive learning, or by imposing a prior on the representation space. Below we discuss these baselines in more detail.

\begin{itemize}
    \item \p{Behavior Cloning (BC)} \cite{bojarski2016end} A standard imitation learning approach. BC learns to map the state to robot actions as demonstrated by the expert. This method does not explicitly encode the state to extract task-relevant information; rather, it attends to the entire state when predicting robot actions.

    \item \p{Variational Autoencoder (VAE)} \cite{kingma2013auto} A self-supervised approach that learns to encode the state into a structured representation space. The decoder is modified to act as the policy head and maps the encoded feature to the expert action. This architecture introduces a bottleneck that tries to force the robot to retain only critical information.

    \item \p{CLASS} \cite{lee2025class} A supervised contrastive learning approach. It aligns observations in the representation space based on the similarity of the future action sequences. This allows the robot to learn shared structure across demonstrations.

    \item \p{VINN} \cite{pari2021surprising} A self-supervised approach that separates representation learning and policy learning. The visual encoder is pretrained using the Bootstrap Your Own Latent (BYOL) \cite{grill2020bootstrap} objective. For new observations, it finds the nearest neighbors in the representation space and chooses the weighted average of the actions corresponding to the neighbors.    
\end{itemize}

\subsection{Simulations}
First, we test our method against baselines in a simulated environment.
The robot is trained on demonstrations collected in one scenario.
We test if \name{} can retain good performance when encountering visual perturbations like changes in the lighting conditions, background variations, etc.

\begin{figure*}
    \centering
    \includegraphics[width=0.9\linewidth]{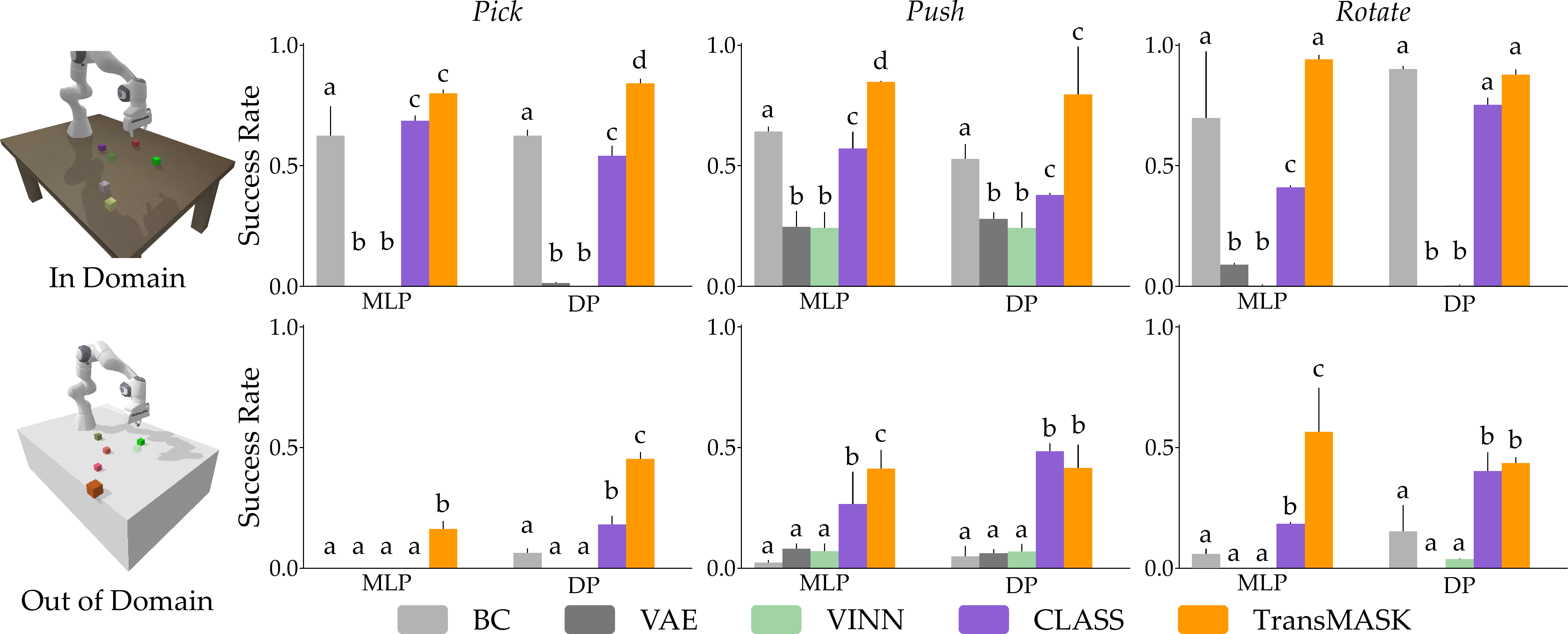}
    \caption{Results from our simulated experiments. We perform the simulated experiments in three tasks --- \textit{Pick}, \textit{Push}, and \textit{Rotate}. The methods are trained with two different policy heads: a fully connected policy (MLP) and a diffusion policy (DP). We evaluate the methods in two scenes, denoted as In-Distribution or ID (over wooden table, top row), and Out-of-Distribution or OOD (over marble table, bottom row). All methods are trained on demonstrations collected in the ID scene, except VINN and CLASS which are trained on demonstratinos from both ID and OOD scenes. We use the Compact Letter Display (CLD) algorithm \cite{piepho2004algorithm} which assigns letters to each policy. The policies which do not share a letter have statistically significant difference in their results.}
    \label{fig:sim_image}
\end{figure*}

\p{Simulation Environment} We train and evaluate all methods on three tasks in the Panda-Gym environment \cite{gallouedec2021panda}.
The environment includes a $7$ degrees of freedom (DOF) Franka Emika robot arm mounted on a table. 
Different colored blocks are scattered in the robot's workspace during data collection and evaluation.
These blocks are randomly initialized within a circle of radius $0.1$ meter in the center of the table.
During evaluation, we test the methods on in-distribution data, where the environment scene is the same as it was during data collection, and on out-of-distribution data with significant visual variations.
The visual perturbations are introduced by changing the position and intensity of the lighting source, changing the table on which the robot is mounted, and changing the appearance of the blocks.
The left columns in \fig{sim_image} display the two scenes.
The figure on top shows the robot setup used for data collection and in-distribution evaluation, while the figure on the bottom shows the setup used for out-of-distribution evaluation.
Demonstrations are collected using a simulated human teacher and we record the ground truth pose of all objects from the simulation environment.

We design three simulated tasks: picking up and placing a block at a target location (\textit{Pick}), pushing a block to a target location (\textit{Push}), and rotating a Rubik's cube to a desired orientation (\textit{Rotate}).
In \textit{Pick} and \textit{Push} tasks, the environment is initialized with a green block that the robot needs to move to a target location. 
This block along with the target location is randomly initialized each time.
Additionally, $4$ different colored blocks are also initialized to act as distractors --- the robot should learn that only the green block and the target location are the relevant features in the state, and therefore the robot can ignore the irrelevant information, i.e., the state of the distractor blocks.
In \textit{Rotate}, the environment is initialized with $5$ Rubik's cube $4$ of which have randomly initialized orientation while the desired cube always starts with the same orientation.
The location of each Rubik's cube is randomly selected at the start of each episode.
The robot must pick and rotate the cube by $90^\circ$ about the longitudinal and vertical axes.
The robot should learn to pick up the target cube with the same orientation while ignoring the other cubes and table color.


\p{Demonstrations} At each timestep of the demonstration we collect an RGB image $I_{env}$ of size $256 \times 256$ from a static camera that observes the entire simulated environment, an ego-centric image $I_{ego}$ captured from a camera mounted on the robot's end-effector of size $256 \times 256$, the robot's state $x$, and the robot's action $a$ applied by the expert human.
In \textit{Pick} and \textit{Push}, $x$ is an $8$-dimensional vector including end-effector pose and a boolean gripper state, while $a$ is a $7$-dimensional vector including linear and angular velocity of the end-effector and the gripper actuation.
In \textit{Rotate}, $x$ is an $8$-dimensional vector including the joint angles of the robot and a boolean gripper state, while $a$ is also an $8$-dimensional vector including the joint velocities and gripper actuation.
This target pose is the Euclidean position where the block should be moved in \textit{Pick} and \textit{Push}; while in \textit{Rotate} it is the desired orientation that the cube must obtain.
Additionally, we extract segmentation masks of all the blocks, the robot, the gripper, and the table in the form of bounding boxes $\{\mathcal{B}\}$.
The coordinates of the bounding boxes are concatenated into a segmentation feature $\phi_{seg}$ of dimension $32$.
This segmentation feature satisfies our assumption that the state is disentangled --- only the bounding boxes of the robot and object are relevant for the task.

\p{Procedure} Since the observations include high-dimensional images, we first use a ResNet-18 vision encoder to map the images $I_{env}, I_{ego}$ to a low-dimensional space $\phi_{env}, \phi_{ego}$.
The state includes these image features, the robot state, and the segmentation feature $s = \left[ \phi_{env}^T, \phi_{ego}^T, \phi_{seg}^T, x^T \right]$.
Our method transforms this state into a state representation $z$ retaining only the relevant information needed to complete the task.
Ideally, $z$ should focus on the robot state and segmentation features corresponding to the target object, i.e., $z = \left[ x, \phi_{seg}' \subset \phi_{seg} \right]$.
For a fair comparison, all the baselines receive the same state $s$ as inputs.

We train the policies on $100$, $100$, and $150$ demonstrations for the \textit{Pick}, \textit{Push}, and \textit{Rotate} tasks, respectively.
We highlight that \name{}, BC, and VAE are trained only on demonstrations collected in the first robot scene shown in the top left corner of \fig{sim_image}.
Since VINN and CLASS are designed to reduce the discrepancy between the image features of different observations with similar behaviors, we train these two baselines on demonstrations that are collected from both the robot scenes.
\textit{Concretely, VINN and CLASS are trained on datasets with half of the demonstrations collected from the first robot scene and half from the second scene.
Our procedure provides an advantage to VINN and CLASS, since they see examples with both types of tables.}

\p{Results}
\fig{sim_image} summarizes the results of our simulated experiments.
All methods are evaluated for $100$ rollouts each in two robot scenes.
In the first scene: In-Distribution (ID), which is shown in the top left corner of \fig{sim_image}, the observations are from the same distribution as the training set.
In the second scene: Out-of-Distribution (OOD), which is shown in the bottom left corner of \fig{sim_image}, there is significant distributional shift in the visual observations as the robot is mounted on a different table and the lighting conditions vary each rollout.
\textbf{Importantly, the observations in OOD are truly out-of-distribution only for \name{}, BC, and VAE, because VINN and CLASS are trained on demonstrations from both these scenes.}
The top row of \fig{sim_image} shows the success rate in ID and the bottom row shows the success rate in OOD.
In each plot, we compare the methods with an MLP action head on the left and a DP action head on the right.

\subsubsection{Can \name{} achieve robustness against visual perturbations?}
In the ID scene, \name{} consistently outperforms the baselines across the three tasks.
This trend holds for both action heads, MLP and DP.
One plausible explanation for the effectiveness of masking even in the ID scene is the presence of distractor objects.
Standard imitation learning policies can form spurious correlations with the clutter in the scene, leading to causal confusion.
Since the target object and distractors are randomly sampled in our setup, the joint configuration of distractors may vary significantly and become out-of-distribution if the policy fails to ignore them.

In OOD scenes, there is significant variation in the lighting conditions --- the location, intensity, and color of the light source varies every episode --- which causes the visual features to experience a prominent distribution shift.
As a result, all policies experience a performance drop of at least $45 \%$.
However, our method is able to significantly outperform the baselines across all the tasks for both MLP and DP policy heads. 
Although there is an exception, the DP policy of \name{} slightly underperforms compared to CLASS in \textit{Push}, a repeated measures ANOVA test found that this difference is not statistically significant.
A likely explanation for why the performance of \name{} in OOD scene almost reduces to half the original performance is the inclusion of ResNet features which violates the assumption of state disentanglement.
While \name{} suppresses the majority of the ResNet features, some elements retain small, non-zero magnitudes.
Since ResNet features are likely to shift drastically with OOD image observations, these residual values negatively affect performance.

\begin{figure}
    \centering
    \includegraphics[width=\linewidth]{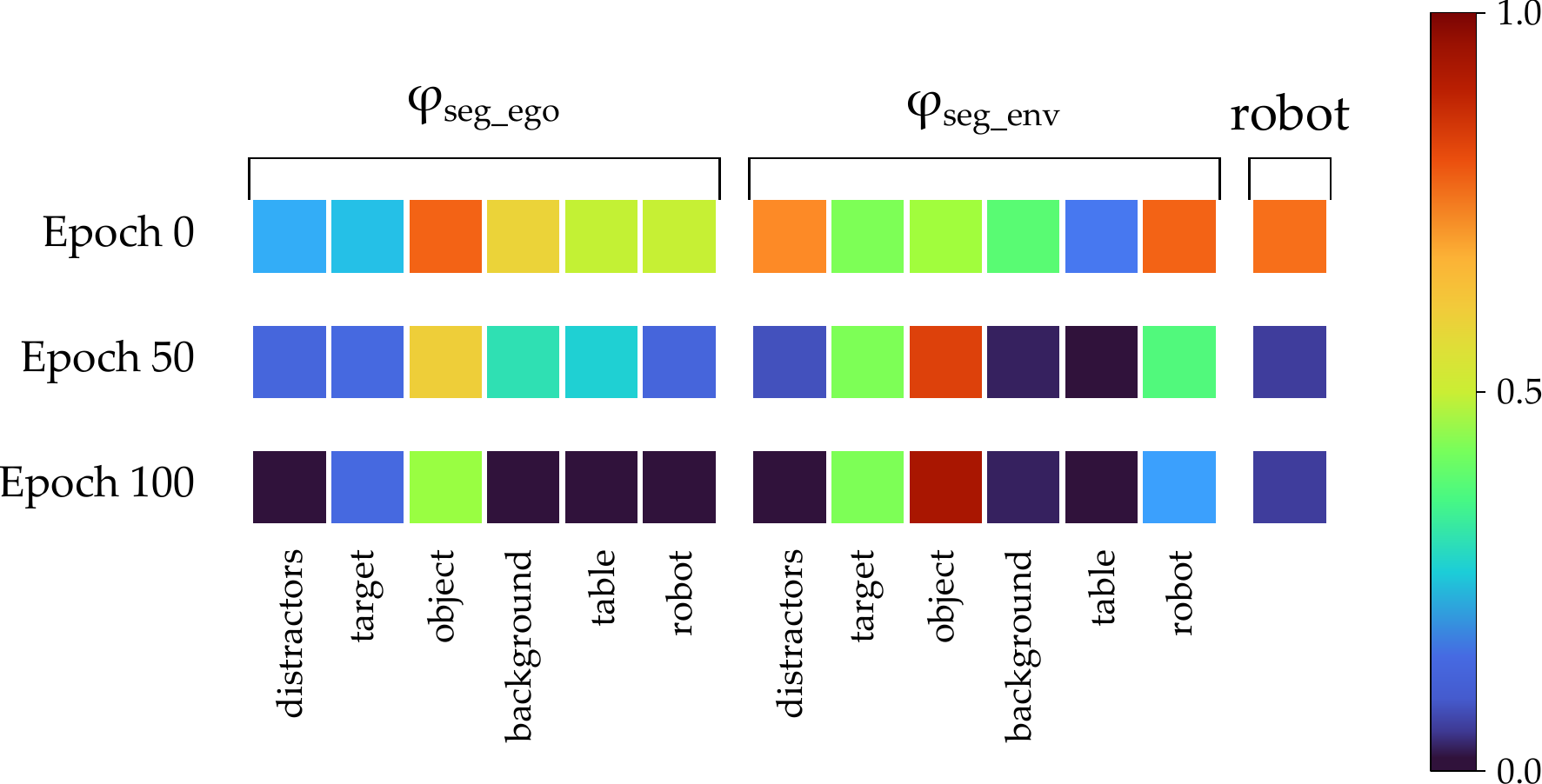}
    \caption{Learned mask over the course of training. Our method learns a mask $M \in \mathbb{R}^{n \times n}$, and computes the latent state as $z = M s$. Therefore, the magnitude of the $i$-th column of $M$ corresponds to the $i^{\text{th}}$ element of $s$. Here, we show how the weights of the state change as the policy training progresses. We label which components of the column relate to which information in the scene. The components with a dark blue shade have values closer to $0$, indicating that the corresponding feature of the scene is removed from $z$. In contrast, the components with a red shade have values closer to $1$, indicating that the associated feature is retained in $z$.}
    \label{fig:mask}
\end{figure}

\subsubsection{Does \name{} learn correct masks?}
After examining the results, we visualize how \name{} learns to weight the elements of the states.
To this end, we visualize the mask $M$ over the course of training in \fig{mask}.
$M$ is an $n \times n$ matrix where $n$ is the dimension of the state.
To visualize how the mask assigns importance to different elements of the state, we normalize each column of $M$.
Specifically, for each column, we compute the sum of all its elements and rescale the resulting column such that its values lie in the range $\left[ 0, 1 \right]$.
These normalized values of the columns indicate which elements of the state are allowed to pass to the state representation $z$.
The normalized values of the columns are close to zero for elements of $s$ that are identified as irrelevant and are not included in $z$.

\fig{mask} shows how the weights of the state change over the course of training. 
In the beginning, since the mask network is initialized randomly, the weights are almost uniform. 
This means that the mask allows all the elements of the state to pass through in the representation $z$.
However, with further training the mask learns to selectively filter out elements that correspond to environment noise.
Concretely, the weights corresponding to the robot proprioceptive states and the bounding boxes of the object, the robot, and the target in the segmentation features $\phi_{seg}$ stay non-zero while those corresponding to the bounding boxes of the table, the robot, and the background are slowly pushed towards zero.
Interestingly, this is found to be true for segmentation features obtained from both the static camera as well as the robot mounted camera.

\begin{figure*}
    \centering
    \includegraphics[width=0.8\linewidth]{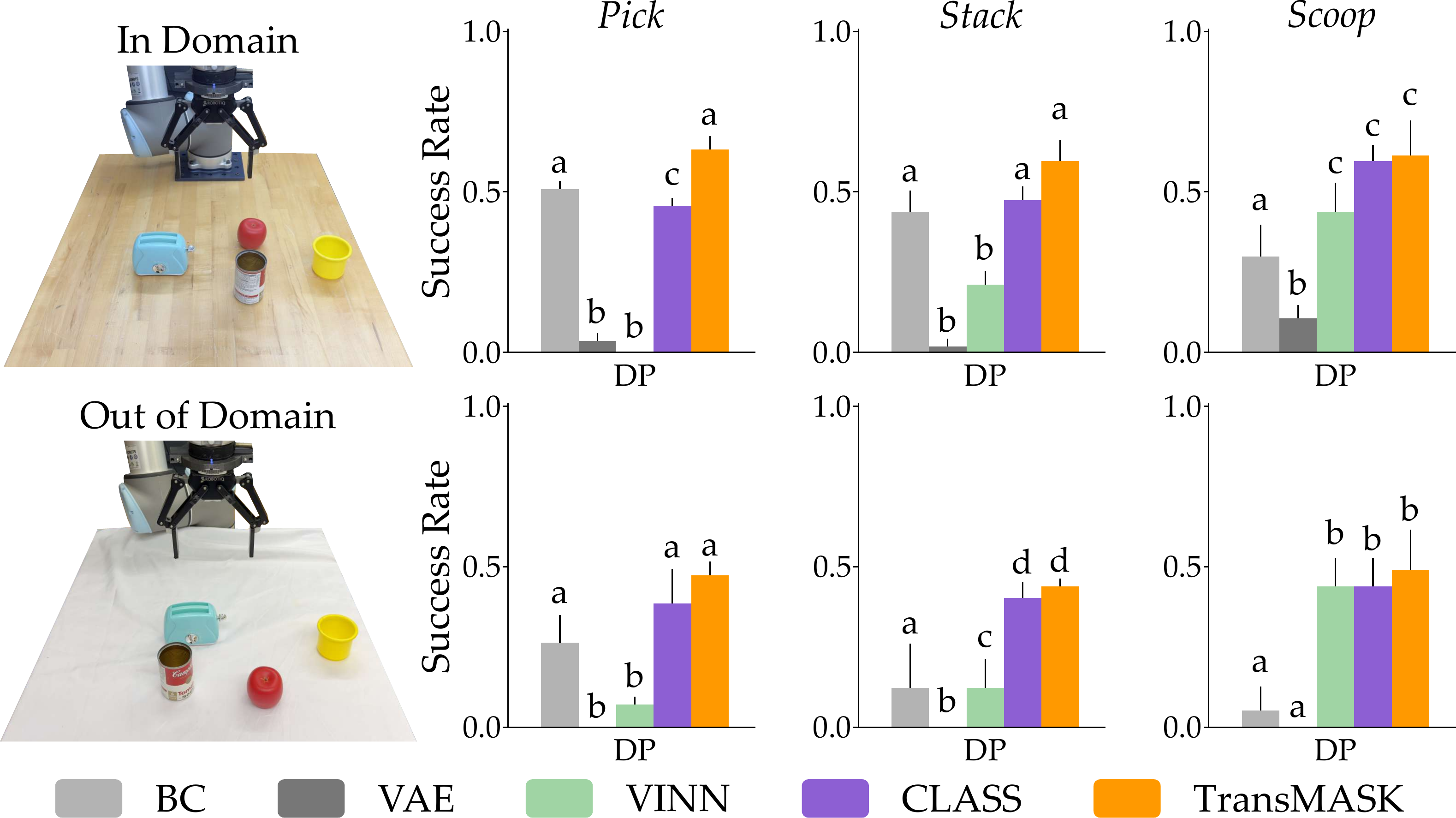}
    \caption{Results of our real-world experiments. We test the methods in three tasks --- \textit{Pick}, \textit{Stack}, and \textit{Scoop}. We evaluate the methods in two scenes, denoted as In-Distribution or ID (over wooden table), Out-of-Distribution or OOD (the table is covered with a white sheet). All the methods are trained on demonstrations collected over the wooden table, except VINN and CLASS which are trained on a mixed dataset with demonstrations collected from ID and OOD scenes. The policies which do not share a letter have statistically significant difference in the results.}
    \label{fig:robot_exp}
\end{figure*}

\subsection{Real-World Experiments}
Next, we move to a real-world setting where a robot arm performs table-top manipulation tasks.
Unlike our simulation environment, the images in a real-world setting present unexpected challenges: the objects have textures and the robot may not segment the objects consistently and accurately.
Our objective is to test if our proposed method can still be effective in this imperfect scenario.

\p{Experiment Setup} We train and evaluate all methods on a $6$ DOF UR10 (Universal Robots) robot mounted on a table.
We use two Logitech C$920$ cameras to capture image observations.
One camera is mounted on the robot's end-effector to record the ego-centric images and the other camera is mounted on the table such that it observes the entire robot environment.
At the start of each task we place the desired object in the environment which the robot needs to manipulate.
Additionally, we place other objects in the environment as distractors that can confuse the robot.
We want to test whether the robot learns to ignore the clutter in the environment and focus only on the object it needs to manipulate.
Similar to the simulated tasks, during evaluation we consider two conditions: in-distribution data, where the scene is the same as during data collection; and out-of-distribution data where we cover the table and the background with a sheet.

We test our method in three tasks: picking up and placing a block at a target location (\textit{Pick}), stacking two LEGO blocks (\textit{Stack}), and scooping from a bowl placed on the table (\textit{Scoop}).
In all tasks, the target object is initialized within a predefined region on the left side of the robot workspace, with $2$-$3$ additional objects randomly placed on the table as distractors.

We compare \name{} against the same baselines used in the simulated experiments --- BC, VAE, VINN, and CLASS.
We train each method on $20$, $25$, and $20$ demonstrations for the \textit{Pick}, \textit{Stack}, and \textit{Scoop} tasks, respectively.
All real-world experiments are conducted using only a diffusion policy (DP).

\p{Demonstrations} We collect robot demonstrations using a joystick to teleoperate the robot.
At each timestep, we collect two images of size $256 \times 256$ from the robot and table mounted cameras ($I_{ego}, I_{env}$ respectively), the robot's state $x$, and the robot action $a$ applied by the human demonstrator.
Robot state $x$ is a $7$-dimensional vector including the end-effector pose and a boolean gripper state, the robot action is also a $7$-dimensional vector consisting of linear and angular end-effector velocity along with the gripper actuation.
To detect the segmentation masks of the objects we use the open-source langSAM model \cite{Medeiros2025LangSegmentAnything} built on SegmentAnything \cite{ravi2024sam2} and GroundingDINO \cite{liu2023grounding}.
langSAM allows us to extract bounding boxes $\{\mathcal{B}\}$ for all the objects in the environment which are concatenated together to form the segmentation feature $\phi_{seg}$.
Notably, while langSAM provides bounding boxes for everything present in the environment, the robot still doesn't explicitly know which bounding box is important for task completion.

\p{Results} \fig{robot_exp} summarizes the results of our real-world experiments.
As in simulated experiments, we consider two robot scenes: first, an In-Distribution (ID) scene where the observations are from the same distribution as the training set, and second, an Out-of-Distribution (OOD) scene in which a distributional shift is introduced by changing the color of the table.
We placed a white sheet on the surface --- note that this also removed reflections and changed the intensity of the shadows.
These ID and OOD scenes are shown in the left column of \fig{robot_exp}.
Each method is evaluated over $20$ rollouts per scene, and we report the success rates averaged over three random seeds.

Across all three tasks, \name{} consistently outperforms all the baselines in the ID scene.
Our method significantly outperforms BC and VAE in the OOD scene as well.
It also substantially outperforms VINN in both scenes on \textit{Pick} and \textit{Stack}.
However, in \textit{Scoop}, \name{} achieves a substantial gain in performance in ID and only a modest improvement in OOD.
Similarly, compared to CLASS our method achieves slight improvement in the success rate in OOD.
Notably, our method is trained exclusively on demonstrations collected in the ID scene, \textbf{whereas VINN and CLASS are trained on a mixed dataset containing equal numbers of ID and OOD demonstrations}.
Although performance in the ID scene is generally higher than in the OOD scene, the degradation under distribution shift is less pronounced than in the simulated experiments.
This is primarily explained by the fact that the target object positions vary less drastically in the real-world setup --- it is sampled from a smaller region, with variations within $5\, \text{cm}$.
\section{Conclusion}
In this manuscript we explored the problem of robust imitation learning under environmental variations.
One way to make learned policies more robust is by pruning the states they are conditioned on; instead of a policy which depends on every observed feature, we want the robot to only reason over features that are relevant to the current task.
To this end, we introduced \name{}, a modular mask which can be added to a variety of standard imitation learning frameworks.
\name{} does not require modifying the imitation learning objective or the training procedure; instead, it relies on the intrinsic gradients of the optimization scheme.
Specifically, \name{} learns a sparse mask matrix that transforms the input state into a compressed representation by eliminating irrelevant components (e.g., background color, table texture, clutter).
The resulting representation preserves features that significantly influence the policy's loss, thereby aligning the learned representations with task structure.
By conditioning the policy on this compressed state representation, \name{} produces policies that generalize better to distributional shifts.

Additionally, we analyzed the commonly used representation strategies in imitation learning, including the information bottleneck objective and contrastive learning.
We showed how these approaches can inadvertently introduce instability or eliminate task-critical information due to the misalignment of the auxiliary objective with policy optimization.
By contrast, \name{} directly exploits gradients from the primary imitation objective, ensuring that the feature selection is tightly coupled to task performance.
The structure of $M$ also prevents \name{} from learning a state embedding which collapses into an action representation.

We tested \name{} through experiments with state-of-the-art baselines in both simulated environments and real-world robotic manipulation tasks.
Across all settings, particularly under environmental perturbation, \name{} consistently achieved significantly higher success rates, demonstrating improved robustness. 

\p{Limitations and Future Work}
Our core limitation lies in our assumption that the input space admits a clear separation between task-relevant and task-irrelevant features, i.e., the state $s$ is sufficiently disentangled into $\mu$ and $\eta$.
In practice, this disentanglement is not guaranteed.
Object-level segmentation offers a promising mechanism to approximate this structure by explicitly decomposing the scene into semantically meaningful features.
Although segmentation models introduce their own failure modes, our real-world experimental results indicate that these do not constitute a significant bottleneck for our proposed method.
A second limitation arises from the implicit learning of the mask.
The mask is not trained with direct supervision; instead, it is optimized through policy gradients during training.
Consequently, its quality depends on the stability of the optimization process and the quality of the training data.
In cases where the loss landscape is noisy or training data is limited, the policy may converge to a suboptimal solution, leading the mask to encode incorrect features.
This issue is particularly likely in small datasets, where irrelevant features may exhibit accidental correlations with actions, causing the mask to include irrelevant features in the state representation.
Therefore, our method should be applied with careful consideration of data quality, dataset size, and optimization stability.

While our experiments focus on imitation learning, the underlying principle is not restricted to this setting.
The approach could potentially be extended to reinforcement learning, particularly in scenarios involving sim-to-real transfer, where robustness to distribution shift and the extraction of task-relevant features are critical.
Future work will focus on exploring the application of \name{} to this application.
Another potential avenue for future research is to provide theoretical guarantee for the convergence of the mask.

\bibliography{bibtex}
\bibliographystyle{unsrt}

\end{document}